\documentclass[english]{article}
\usepackage{spconf,amsmath,graphicx}
\usepackage[T1]{fontenc}
\usepackage[latin9]{inputenc}
\usepackage{textcomp}
\usepackage{graphicx}
\usepackage{babel}
\usepackage{multirow}
\usepackage{xcolor}
\definecolor{ao}{rgb}{0.0, 0.5, 0.0} 
\usepackage{tabularx}

\begin{document}

% Title.
% ------
\title{Fine-grained Wound Tissue Analysis using Deep Neural Network}
%
% Single address.
% ---------------
%\name{Yiren Zhou}
%\address{Author Affiliation(s)}
%
% For example:
% ------------
%\address{School\\
%	Department\\
%	Address}
%
% Two addresses (uncomment and modify for two-address case).
% ----------------------------------------------------------
\name{
H. Nejati, 
H. A. Ghazijahani,
M. Abdollahzadeh,
T. Malekzadeh,
N.-M. Cheung,
K.-H. Lee, L.-L. Low
}
\address{Singapore University of Technology and Design (SUTD), Singapore 487372}
  %{K.H. Lee, L.L. Low, L.F. Liew}
	%{Bright Vision Hospital}
% \twoauthors
%   {Yiren Zhou, Sahar Iravani, Hossein Nejati, Ngai-Man Cheung}
% 	{Singapore University of Technology and Design}
%   {Kheng-Hock Lee, Lian-Leng Low, Joo-Ming Ng, Lee-Foong Liew}
% 	{Bright Vision Hospital}

\maketitle

\begin{abstract}
	Tissue assessment for chronic wounds is the basis of wound grading and selection of treatment approaches. While several image processing approaches have been proposed for automatic wound tissue analysis, there has been a shortcoming in these approaches for clinical practices.  In particular, seemingly, all previous approaches have assumed only 3 tissue types in the chronic wounds, while these wounds commonly exhibit 7 distinct tissue types that presence of each one changes the treatment procedure. In this paper, for the first time, we investigate the classification of 7 wound tissue types.  
We work with wound professionals to build a new database of 7 types of wound tissue.    
We propose to use pre-trained deep neural networks for feature extraction and classification at the patch-level. We perform experiments to demonstrate that our approach outperforms other state-of-the-art. We will make our database publicly available  to facilitate research in wound assessment.
\end{abstract}


\section{Introduction}

Chronic wounds are a major threat to public health and economy. They
are the byproduct of the frailty associated with either aging or
diabetic patients, with a growing number worldwide \cite{ConstantineBillsLaveryEtAl2014}. These wounds require frequent
visits to hospital and do not heal for months and often years, and if left open, the patient is increasingly subject
to risk of infection, amputation and even death. On the other hand, 
the healthcare cost to provide properly and personalized care to these patients is enormous. Therefore, there is a pressing need for automatic approaches to aid caregivers and medical personnel. 

The first step for wound treatment is wound grading, in which medics describe the wound by its dimensions and the color of its tissue composition. There are 7 tissue types commonly present at
the wound site \cite{SenGordilloRoyEtAl2009}: necrotic, sloughy, healthy granulating, unhealthy granulating, hyper granulating, infected, and epithelizing. {\em Necrotic} is the dead tissue and is black
in color. It occurs when skin cells inside of the wound die off. The {\em sloughy} tissue is a type of wet necrotic
tissue that is detaching itself from the wound site, and is often seen white,
yellow or grey in color. {\em Healthy granulating} is the new grown tissue that is generated when the wound surface area
is starting to heal by tiny blood vessels that appear at the surface, with light red
or pink in color, and will be moist. {\em Unhealthy granulating} tissue is when the process of granulation is irritated
by problems such as infection or lack of good blood supply, and appears dark red, bluish, or very pale, and may indicate ischemia or infection in the wound. {\em    Hyper granulating} tissue is the tissue that grows above the wound margin when the proliferative phase of healing is prolonged usually as a result of bacterial imbalance or irritant forces. {\em Infected} tissue is greenish color tissue with foul smell caused by bacterial infection that may spread to different parts of the wound and it surrounding tissues. Finally, {\em epithelizing} tissue is a group
of tightly-packed cells that provides protective layers over the granulating
tissue.

\begin{figure*}[t]
	\begin{centering}
		\includegraphics[width=.9\textwidth]{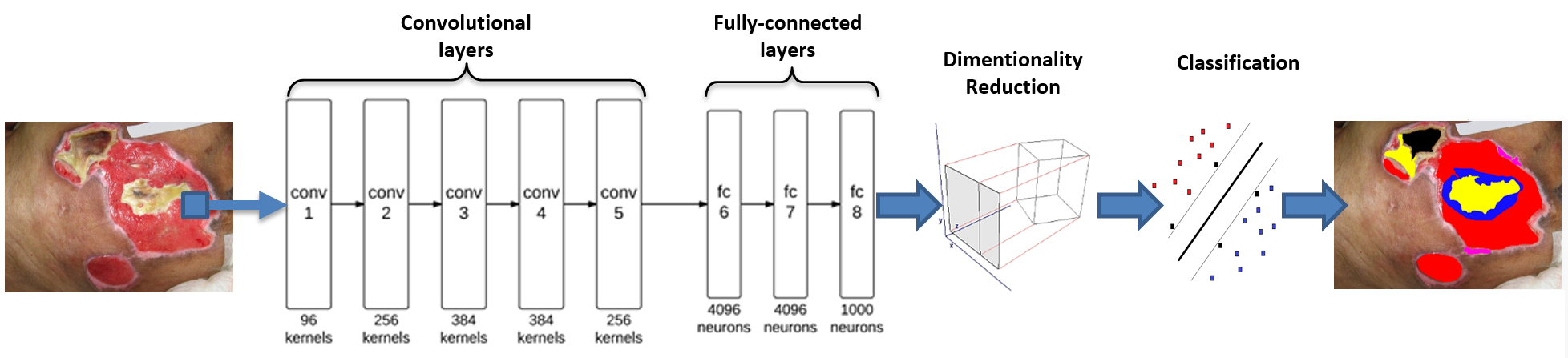}
		\par\end{centering}
	\caption{\label{fig:BlockDiagram}Tissue classification block diagram: image
		patches centered around each pixel is fed to DNN. The fully connected
		layers of DNN are treated as features, which are then subjected to
		dimensionality reduction and classification.}
\end{figure*}

Several automatic wound tissue classification approaches have been proposed in the literature, such as \cite{WannousTreuilletLucas2007,WannousTreuilletLucas2010,WangPedersenStrongEtAl2013}.
As the first step, wound area is selected using either automatic (e.g. in \cite{WangPedersenStrongEtAl2015}) or semi-automatic (e.g. in \cite{Oduncu2005}) techniques. Following is usually the image pre-processing step for color correction and white balance estimation (e.g. in \cite{Haeghen2000}). Tissue classification step is then performed, by incorporating one or several image descriptors and classification. The most commonly used features are color histograms (e.g. in \cite{Hani2011}, texture parameters such as entropy, sum of squares variance, wavelet, and local binary patterns (LBP) (e.g. in \cite{Noguchi2014}). While there are differences in the requirements and robustness of these approaches, an important assumption in seemingly all of these approaches has  undermined their usability. These methods assume that there are \emph{only 3 tissue types} (Necrotic, Sloughy, and Granulation) present at the wound bed, ignoring and combining other types. This is while in modern medical practices chronic wound tissues are categorized into the aforementioned 7 types, with each one affecting the treatment procedures. Clustering the real 7 tissue types into 3 clusters can therefore be insufficient for clinical use.

In this work, we propose an automatic wound tissue classification system that correlates to actual clinical assessment and supports clinical decision making. Working with wound professionals, we firstly collected a dataset of chronic wounds and labeled into 7 types. We propose to use layers of a pre-trained deep neural network (DNN) as high-level image representations, and subject them to dimensionality reduction. This smaller set of features is then used to train an SVM classifier to label the wound image into 7 tissue types. For our experiments we use AlexNet \cite{AlexNet2012} trained on LSVRC-2010 ImageNet training set  \cite{ImageNet2010}. Our results on 350 clinically assessed chronic wound images and comparison with previous approaches show accurate and robust classification of 7 tissue types.
Our contributions in this work are included: (I) address the fine-grained, clinically-relevant wound tissue classification problem of 7 tissue types. To the best of our knowledge, this is the first attempt to classify more than 4 wound tissue types.  (II)  propose an accurate and robust wound tissue analysis using DNN model transfer.  (III) We will make available an image dataset with clinically approved labeling.  Labeled dataset for wound is scarce and requires tremendous effort to build, but is important for wound assessment research. (IV) We solve an NP-hard optimization based on \textit{Knapsack} problem to reach a balanced distribution of tissue types in both train and test sets.

\section{\label{sec:Methodology}Methodology}

We propose to use a supervised Deep Neural Network to determine the tissue types \cite{pomponiu2016deepmole}. While training a DNN from scratch requires a significantly large training dataset, recent works have shown that the higher layers of a DNN trained on a large labeled dataset could be general enough for another image classification task (a.k.a. transfer learning) \cite{yosinski2014transferable}. We here propose to reuse a pre-trained DNN as a feature extractor, instead of using it directly as classifier.

We present our classification pipeline as follows: Each image is labeled based on the included tissue types and is partitioned into $n \times n$ patches. The class of each patch is determined based on the majorities of included pixels. This square patch is then fed to the DNN as an input. Next, instead of using DNN classification output, we treat DNN layers as image features. In other words, we rely on layers of the DNN to extract the high-level information as image representations in a high-dimensional space. We then apply dimensionality reduction and training on these features to reach the final patch label (here performed with an SVM classifier). In this work, we consider AlexNet \cite{AlexNet2012} as our DNN, and use Matconvnet \cite{vedaldi2015matconvnet}, a widely-adopted open source deep learning framework. Figure \ref{fig:BlockDiagram} illustrates the block diagram of our proposed approach. As illustrated in Figure \ref{fig:BlockDiagram}, AlexNet structure has 5 convolutional layers ($conv1$ to $conv5$) and 3 fully-connected layers ($fc6$, $fc7$ and $fc8$). Each convolutional layer contains multiple kernels, and each kernel represents a 3-D filter connected to the outputs of the previous layer. Each fully-connected layer contains multiple neurons that each one is connected to all the neurons in the previous layer. The weight of each connection is optimized during the original training on the ImageNet dataset. Different layers in a DNN are often considered to have different level of features. The first few layers contain general features that resemble Gabor filters or blob features. The higher layers contain specific features, each representing particular class in dataset \cite{yosinski2014transferable}. Thus features in higher layers are considered to have higher level information compared to general features in base layers.

Employing transfer learning using AlexNet, we need to consider two main factors, namely, the size of the new dataset, and the similarity between the original and the new datasets \cite{yosinski2014transferable}. AlexNet model is trained on the ILSVRC-2012 dataset with 1.2 million images in 1000 categories, including general kinds of natural and man-made images \cite{ImageNet2010}. We here intent to use this model to classify our dataset of wound image patches that is significantly smaller compare to original ILSVRC-2012 dataset. It is therefore highly likely that fine-tuning AlexNet on our wound image dataset would result in an over-fitted model, and thus we use AlexNet as a fixed feature extractor instead.

The second concern is the difference between the nature of the tissue image classification, and the image classification task AlexNet originally trained for. Despite this difference in the classification task, previous works such as \cite{yosinski2014transferable,razavian2014cnn,zhou2016vehicle} have reported the fully connected layers to contain high-level information, seemingly much wider than what is needed for the original classification task. In order to examine this hypothesis and find the best feature set to fulfill our purpose, we assess all three fully connected layers (referred to by $fc6$, $fc7$, and $fc8$ in Figure \ref{fig:BlockDiagram}) in AlexNet for their discriminative power in wound tissue classification. We do not consider the convolutional layers due to their sizes (43264 features in the smallest) that are too large for our current dataset.

For each image patch extracted from the wound image, we resize it to $227\times227$, make it valid AlexNet input. We extract the fully connected layers as image representations, namely, $fc6$, $fc7$, and $fc8$, with 4096-, 4096-, and 1000-dimension vectors respectively. We then apply Principal Component Analysis (PCA) on the extracted layer feature vector, $f=[f_1,f_2,...,f_{4096}]$, to reduce $f$ to a vector $f0= [f0_1,f0_2,...,f0_m]$, with $m=18$ dimensionality. The resulting DNN-based feature vectors, $f0$, are then used to train a linear SVM classifier.
SVM is trained using k-fold cross validation. 
It is important to have  even distribution of data set between folds with respect to the tissue class types. This problem can be formulated as  an NP-hard \textit{Knapsack problem}, and we solve this  with a greedy approach, to reach a balanced distribution of tissue types in both train and test sets. Specifically, 
in each step, the fold of one image is determined by solving the Knapsack problem. The cost is defined as the standard deviation (SD) of tissue types distribution over folds in each step. Considering $l_{n,c}$ as the number of class $c$ patches in $n$-th image , the total number of class $c$ patches in fold $k$, in step $s$ is defined as:

\noindent
\begin{equation}
L_{k,c,s} = \sum_{n=1}^{\lambda_{k,s}}l_{n,c}
\label{eq1}
\end{equation}
where, $\lambda_{k,s}$ is the number of images placed in fold $k$ till step $s$. The mean value of total number of class $c$ patches in $s$-th step is:
\begin{equation}
\overline{L_{c,s}} = \dfrac{1}{k}\sum_{k=1}^{K}L_{k,c,s}
\label{eq2}
\end{equation}
The SD of class $c$ patches in step $s$ is calculated as:

\begin{equation}
\sigma_{c,s} = \sqrt{\dfrac{\sum_{k=1}^{K}\left( L_{k,c,s}-\overline{L_{c,s}}\right) ^2}{K}}
\label{eq3}
\end{equation}
The total SD is the sum of SDs over all classes, $\sigma_{s} = \sum_{c=1}^{C}\sigma_{c,s}$,
which is used as the cost function in our optimization problem. In each step, the fold of related image is determined in a way to minimize the total cost.
Note that since we focus on the wound tissue classification, we do not consider pre-processing steps such as wound area detection, and proceed with an already selected Region of Interest (ROI) of the wound. Chronic wound area detection is well-studied in previous works such as \cite{WangPedersenStrongEtAl2015}. 

%\begin{equation}
%\sigma_{s} = \sum_{c=1}^{C}\sigma_{c,s}
%\label{eq4}
%\end{equation}
\begin{figure}[t]
	\begin{centering}
		\begin{tabular}{cc}
			\multicolumn{2}{c}{\textbf{fc8}}\tabularnewline
			{\small{}Excit.} & \includegraphics[width=0.7\columnwidth]{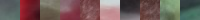}\tabularnewline
			{\small{}Inhib.} & \includegraphics[width=0.7\columnwidth]{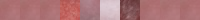}\tabularnewline
			\multicolumn{2}{c}{\textbf{RGB}}\tabularnewline
			{\small{}Excit.} & \includegraphics[width=0.7\columnwidth]{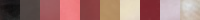}\tabularnewline 
			{\small{}Inhib.} & \includegraphics[width=0.7\columnwidth]{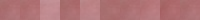}\tabularnewline
			\multicolumn{2}{c}{\textbf{HSV}}\tabularnewline
			{\small{}Excit.} & \includegraphics[width=0.7\columnwidth]{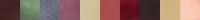}\tabularnewline
			{\small{}Inhib.} & \includegraphics[width=0.7\columnwidth]{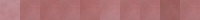}\tabularnewline
			\multicolumn{2}{c}{\textbf{LBP}}\tabularnewline
			{\small{}Excit.} & \includegraphics[width=0.7\columnwidth]{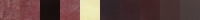}\tabularnewline 
			{\small{}Inhib.} & \includegraphics[width=0.7\columnwidth]{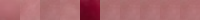}\tabularnewline
		\end{tabular}
		\par\end{centering}	
	\caption{\label{fig:MeanImages}Mean image patches that cause excitation and
		inhibition in 10 highest contributing features.}
\end{figure}

\begin{table*}[]
	\centering
	\caption{Accuracy of different methods versus tissue types.}
	\label{Tab:tbl.1}
	\begin{tabular}{|c||c|c|c|c|c|c|c|c|}
		\hline
		{\small{}Tissue Type} & {\small{}Necrotic} & {\small{}Healthy Gran.} & {\small{}Slough} & {\small{}Infected} & {\small{}Unhealthy Gran.} & {\small{}Hyper Gran.} & {\small{}Epithelialization} & {\small{}Overall} \\ \hline\hline
		AlexNet & 90.65 & 83.12 & 80.88 & 95.54 & 82.10 & 94.17 & 78.34 & 86.40 \\ \hline
		HSV & 75.16 & 83.62 & 85.70 & 87.87 & 65.20 & 75.73 & 69.70 & 77.57 \\ \hline
		LBP & 82.94 & 85.42 & 82.98 & 89.93 & 83.61 & 80.81 & 51.93 & 79.66 \\ \hline
        HSV+LBP & 77.75 & 80.89 & 82.96 & 77.33 & 80.41 & 82.61 & 57.69 & 77.09 \\ \hline
	\end{tabular}
\end{table*}

\begin{figure}[h]
	\begin{centering}
		\includegraphics[width=1\columnwidth]{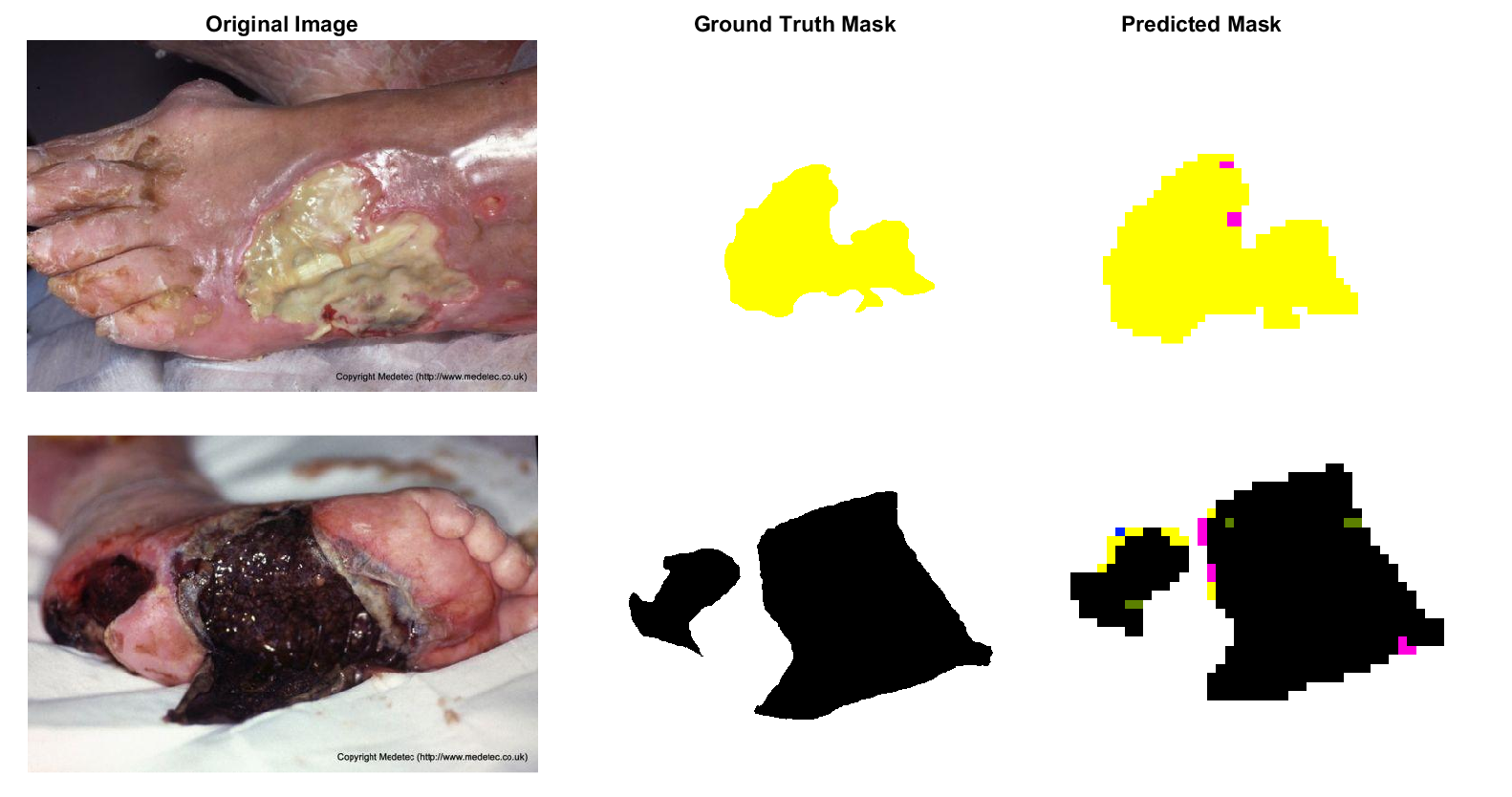}
		\par\end{centering}
	\caption{\label{fig:example}Examples of tissue labeling. From left to right: wound image, ground truth and our method.}
\end{figure}

\begin{figure}[h]
	\begin{centering}
		\includegraphics[width=.9\columnwidth]{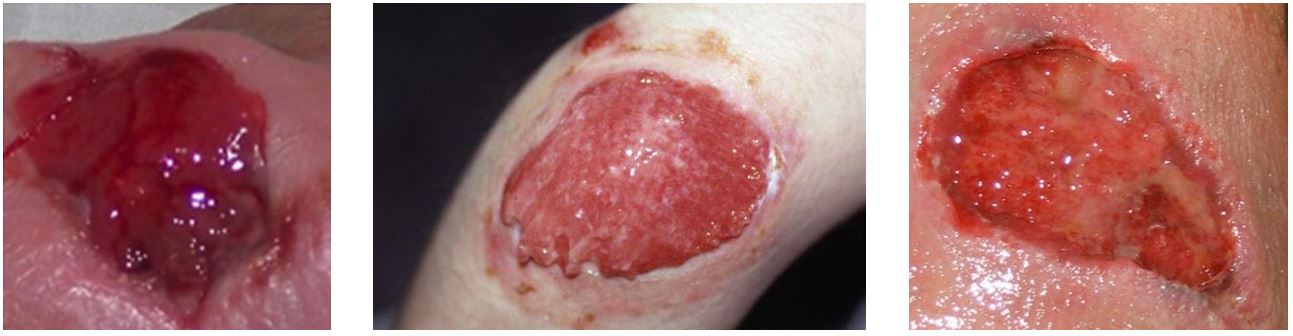}
		\par\end{centering}
	\caption{\label{fig:bad images}sample images with large errors. Tissue types from left to right: unhealthy-, hyper-, and healthy-granulation.}
\end{figure}

\begin{figure}[h]
	\begin{centering}
		\includegraphics[width=.9\columnwidth]{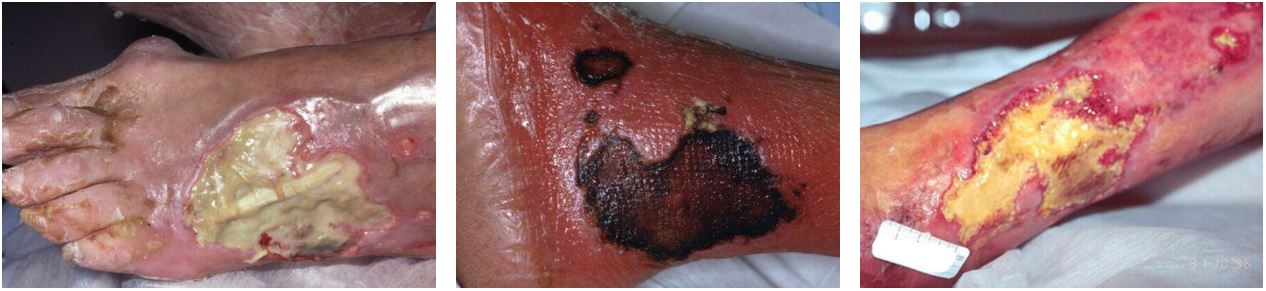}
		\par\end{centering}
	\caption{\label{fig:good images}sample images with small errors. Tissue types from left to right: infected, necrotic and slough.}
\end{figure}

\section{Experimental Evaluations}

In this section we present and discuss performance measures of our method.

\emph{Dataset:}
Our dataset of wound tissues consists of 350 images of chronic wounds, captured in different conditions (illumination, pose, etc.), with different camera devices, with different resolutions (ranging from $216\times158$ to $4208\times2368$). While the majority of these images are collected by our team, for the sake of diversity we added a subset of low-resolution images from the web \cite{mukherjee2014automated}. 
%All the patients participating in this study have been notified about the purpose of the study and completed the related consent form. All the images have been checked for identifiable information and corrected if found. 
Working with wound care specialists, we manually label all images based on clinical wound assessment procedure guidelines. As we will be providing pixel-level labels for each image, it is important to note that the top 3 tissue types in terms of number of labeled pixels are sloughy, necrotic, and then healthy granulating. This uneven distribution is due to the distribution of collaborating patients, which is being addressed in our next data collection. 

\emph{Experimental Setup:}
In this work, we propose a patch-based scheme for wound tissue classification.
We partition the images into $20 \times 20$ patches and classify each wound patch into one tissue class.\\
The classification is a two-step process: First, we compute a set of features for each patch; second, we build a
classification model based on the extracted features.
To build a set of discriminative features, we run AlextNet and extract fc6, fc7, and
fc8 layers output and then apply PCA on the extracted features.
To match the input size of AlexNet, each patch is then resized to $227 \times 227 \times 3$.
For the classification step, we use SVM with a linear kernel. 
In the experiment, we split the data into disjoint training
and testing sets, in a manner that the data which is present in
the training set is not allowed to be in the testing set. 
But in order to make these two sets completely disjoint, we employ k-fold cross validation on the images rather than the patches, i.e. the patches of a particular image have the same cross validation index as their parent image.
This approach prevents having highly correlated data in both training and
testing sets, improving the 
generalizability
of the results.

\emph{Results and discussions:}
Table \ref{Tab:tbl.1} reports the classification accuracy for seven wound tissue types using extracted features from pre-trained DNN and conventional features. As mentioned before, we use 3-fold cross validation. Correspondingly, the mean values of all three folds' results are reported in this table.\\
We have used \emph{AlexNet} as pre-trained network for feature extraction. Besides, in order to compare the performance of classification,we have used the RGB and HSV histograms as color descriptors and the LBP as a texture descriptor. Color and texture features were fed to the classifier. As one can see, using pre-trained DNN as feature extractor, results in better classification accuracy compared to conventional features. In our previous work \cite{nejati2016smartphone}, we have shown that the conventional features revealed a high discriminative power in the three-class scenario. However, they failed to reach an acceptable level of performance when tested in the seven-class scenario and thus cannot be used for clinical purposes. While in three class scenarios each class can be separated using simple features like color and texture, in realistic seven class tissue types more powerful features are needed. We have also analyzed the results and extracted the images with large error, which degrade the overall evaluation parameters. Figure \ref{fig:bad images} shows three of such images. Besides, there are some images that have very small error. Some of these images that have small error are shown in Figure \ref{fig:good images}. Also, Figure \ref{fig:example} illustrates some examples of patch level prediction of different tissue types by our algorithm.
Furthermore, we investigate the mean image representation of patches that excite or inhibit highest contributing dimensions in each feature space. Excitation(/inhibition) mean image is calculated by averaging all patches that lead to top(/bottom) values for each feature dimension.
This comparison illustrated in Figure \ref{fig:MeanImages} shows that (DNN-based features) assign a low value to skin-like patches. On the other hand, these features respond to a variety of different colors and textures that may represent different tissue types. This suggests that DNN layers can extract features of different nature, including color, edge, and texture. It seems therefore, that DNN-based features represent patches in a feature space that not only includes traditional color/texture, but also additional higher level information that led to their better discriminating power.

\section{Conclusions}

In this work we shed light on 
fine-grained tissue classification to better 
%the incorrect assumption of previous works on the tissue types present in chronic wounds, and argued to 
realign the goal with clinically approved practices. We then presented our approach to classify all 7 different tissue types, based on using a pre-trained DNN as a feature extractor for wound tissue classification. We used DNN layers as image representation features and then perform feature reduction and classification using PCA and linear SVM, to reach patch-level labeling of the wound image. In our experiments, we showed that the proposed method not only outperforms previously proposed features, it is more robust in discrimination of similar looking tissue types and also against illumination condition changes. We will make our current dataset publicly available. In our future steps, we will investigate  classification on smart-phones for an accessible solution, and address the associated technical challenges \cite{zhou:tcsvt:2016, zhou2017blur}.

\bibliographystyle{IEEEBib}
{\small{}
\bibliography{WoundBib}

\begin{thebibliography}{10}

\bibitem{ConstantineBillsLaveryEtAl2014}
Ryan~S Constantine, Jessica~D Bills, Lawrence~A Lavery, and Kathryn~E Davis,
\newblock ``Validation of a laser-assisted wound measurement device in a wound
  healing model,''
\newblock {\em International Wound Journal}, pp. n/a--n/a, 2014.

\bibitem{SenGordilloRoyEtAl2009}
Chandan~K Sen, Gayle~M Gordillo, Sashwati Roy, Robert Kirsner, Lynn Lambert,
  Thomas~K Hunt, Finn Gottrup, Geoffrey~C Gurtner, and Michael~T Longaker,
\newblock ``Human skin wounds: a major and snowballing threat to public health
  and the economy.,''
\newblock {\em Wound Repair Regen}, vol. 17, no. 6, pp. 763--71, 2009.

\bibitem{WannousTreuilletLucas2007}
Hazem Wannous, Sylvie Treuillet, and Yves Lucas,
\newblock ``Supervised tissue classification from color images for a complete
  wound assessment tool,''
\newblock in {\em Engineering in Medicine and Biology Society, 2007. EMBS 2007.
  29th Annual International Conference of the IEEE}. IEEE, 2007, pp.
  6031--6034.

\bibitem{WannousTreuilletLucas2010}
Hazem Wannous, Sylvie Treuillet, and Yves Lucas,
\newblock ``Robust tissue classification for reproducible wound assessment in
  telemedicin environments,''
\newblock {\em Journal of Electronic Imaging}, vol. 19, no. 2, pp. 23002, 2010.

\bibitem{WangPedersenStrongEtAl2013}
Lei Wang, Peder~C Pedersen, Diane Strong, Bengisu Tulu, and Emmanuel Agu,
\newblock ``Wound image analysis system for diabetics,''
\newblock in {\em SPIE Medical Imaging}. International Society for Optics and
  Photonics, 2013, pp. 866924--866924.

\bibitem{WangPedersenStrongEtAl2015}
Lei Wang, P.C. Pedersen, D.M. Strong, B.~Tulu, E.~Agu, and R.~Ignotz,
\newblock ``Smartphone-based wound assessment system for patients with
  diabetes,''
\newblock {\em Biomedical Engineering, IEEE Transactions on}, vol. 62, no. 2,
  pp. 477--488, Feb 2015.

\bibitem{Oduncu2005}
H.~Oduncu, V.~Aslanta, M.~Tunckanat, and R.~Kurban,
\newblock ``Skin wound analysis using digital image processing,''
\newblock in {\em Signal Processing and Communications Applications Conference,
  2005. Proceedings of the IEEE 13th}, May 2005, pp. 645--648.

\bibitem{Haeghen2000}
Y.V. Haeghen, J.M.A.D. Naeyaert, I.~Lemahieu, and W.~Philips,
\newblock ``An imaging system with calibrated color image acquisition for use
  in dermatology,''
\newblock {\em Medical Imaging, IEEE Transactions on}, vol. 19, no. 7, pp.
  722--730, July 2000.

\bibitem{Hani2011}
A.F.M. Hani, L.~Arshad, A.S. Malik, A.~Jamil, and F.Y.B. Bin,
\newblock ``Assessment of chronic ulcers using digital imaging,''
\newblock in {\em National Postgraduate Conference (NPC), 2011}, Sept 2011, pp.
  1--5.

\bibitem{Noguchi2014}
H.~Noguchi, A.~Kitamura, M.~Yoshida, T.~Minematsu, T.~Mori, and H.~Sanada,
\newblock ``Clustering and classification of local image of wound blotting for
  assessment of pressure ulcer,''
\newblock in {\em World Automation Congress (WAC), 2014}, Aug 2014, pp.
  427--432.

\bibitem{AlexNet2012}
Alex Krizhevsky, Ilya Sutskever, and Geoffrey~E. Hinton,
\newblock ``Imagenet classification with deep convolutional neural networks,''
\newblock in {\em Advances in Neural Information Processing Systems 25},
  F.~Pereira, C.J.C. Burges, L.~Bottou, and K.Q. Weinberger, Eds., pp.
  1097--1105. Curran Associates, Inc., 2012.

\bibitem{ImageNet2010}
Olga Russakovsky, Jia Deng, Hao Su, Jonathan Krause, Sanjeev Satheesh, Sean Ma,
  Zhiheng Huang, Andrej Karpathy, Aditya Khosla, Michael Bernstein,
  Alexander~C. Berg, and Li~Fei-Fei,
\newblock ``{ImageNet Large Scale Visual Recognition Challenge},''
\newblock {\em International Journal of Computer Vision (IJCV)}, vol. 115, no.
  3, pp. 211--252, 2015.

\bibitem{pomponiu2016deepmole}
Victor Pomponiu, Hossein Nejati, and N-M Cheung,
\newblock ``Deepmole: Deep neural networks for skin mole lesion
  classification,''
\newblock in {\em Image Processing (ICIP), 2016 IEEE International Conference
  on}. IEEE, 2016, pp. 2623--2627.

\bibitem{yosinski2014transferable}
Jason Yosinski, Jeff Clune, Yoshua Bengio, and Hod Lipson,
\newblock ``How transferable are features in deep neural networks?,''
\newblock in {\em Advances in Neural Information Processing Systems}, 2014, pp.
  3320--3328.

\bibitem{vedaldi2015matconvnet}
Andrea Vedaldi and Karel Lenc,
\newblock ``Matconvnet: Convolutional neural networks for matlab,''
\newblock in {\em Proceedings of the 23rd ACM international conference on
  Multimedia}. ACM, 2015, pp. 689--692.

\bibitem{razavian2014cnn}
Ali~S Razavian, Hossein Azizpour, Josephine Sullivan, and Stefan Carlsson,
\newblock ``Cnn features off-the-shelf: an astounding baseline for
  recognition,''
\newblock in {\em 2014 IEEE Conference on Computer Vision and Pattern
  Recognition Workshops (CVPRW)}. IEEE, 2014, pp. 512--519.

\bibitem{zhou2016vehicle}
Y~Zhou, H~Nejati, TT~Do, NM~Cheung, and L~Cheah,
\newblock ``Image-based vehicle analysis using deep neural network: A
  systematic study,''
\newblock in {\em Proc. IEEE International Conference on Digital Signal
  Processing (DSP)}, 2016.

\bibitem{mukherjee2014automated}
Rashmi Mukherjee, Dhiraj~Dhane Manohar, Dev~Kumar Das, Arun Achar, Analava
  Mitra, and Chandan Chakraborty,
\newblock ``Automated tissue classification framework for reproducible chronic
  wound assessment,''
\newblock {\em BioMed research international}, vol. 2014, 2014.

\bibitem{nejati2016smartphone}
Hossein Nejati, Victor Pomponiu, Thanh-Toan Do, Yiren Zhou, Sahar Iravani, and
  Ngai-Man Cheung,
\newblock ``Smartphone and mobile image processing for assisted living:
  Health-monitoring apps powered by advanced mobile imaging algorithms,''
\newblock {\em IEEE Signal Processing Magazine}, vol. 33, no. 4, pp. 30--48,
  2016.

\bibitem{zhou:tcsvt:2016}
Y~Zhou, TT~Do, H~Zheng, NM~Cheung, and L~Fang,
\newblock ``Computation and memory efficient image segmentation,''
\newblock {\em IEEE Transactions on Circuits and Systems for Video Technology},
  2016.

\bibitem{zhou2017blur}
Y~Zhou, S~Song, and NM~Cheung,
\newblock ``On classification of distorted images with deep convolutional
  neural networks,''
\newblock in {\em Proc. IEEE ICASSP}, 2017.

\end{thebibliography}
}
\end{document}